\def\year{2020}\relax
\documentclass[letterpaper]{article} 
\usepackage{aaai20}  
\usepackage{times}  
\usepackage{helvet} 
\usepackage{courier}  
\usepackage[hyphens]{url}  
\usepackage{graphicx} 
\usepackage{color}
\usepackage{subfigure}
\usepackage{amsmath}
\usepackage{amssymb}
\usepackage{bm}
\usepackage[ruled,linesnumbered,noend]{algorithm2e}
\usepackage{algorithmic}
\usepackage{array}
\usepackage{graphicx}  
\title{Object Instance Mining for Weakly Supervised Object Detection}
\author{\Large \textbf{Chenhao Lin$^1$, Siwen Wang$^2$\thanks{Work performed as an intern in SenseTime Research \newline \indent \ \  https://github.com/bigvideoresearch/OIM \newline \indent \ \ $^\dagger$Corresponding author}, Dongqi Xu$^1$$^*$, Yu Lu$^1$$^\dagger$, Wayne Zhang$^1$}\\ 
\textsuperscript{\rm 1}SenseTime Research\\ \textsuperscript{\rm 2}Dalian University of Technology, Dalian, China, 116024\\ 
linchenhao@sensetime.com, wangsiwendut@gmail.com, isdongqixu@gmail.com\\ luyu@sensetime.com, wayne.zhang@sensetime.com 
}

\makeatletter
\def\@copyrightspace{\relax}
\makeatother

\begin{document}

\maketitle

\begin{abstract}
Weakly supervised object detection (WSOD) using only image-level annotations has attracted growing attention over the past few years. Existing approaches using multiple instance learning easily fall into local optima, because such mechanism tends to learn from the most discriminative object in an image for each category. Therefore, these methods suffer from missing object instances which degrade the performance of WSOD. To address this problem, this paper introduces an end-to-end object instance mining (OIM) framework for weakly supervised object detection. OIM attempts to detect all possible object instances existing in each image by introducing information propagation on the spatial and appearance graphs, without any additional annotations. During the iterative learning process, the less discriminative object instances from the same class can be gradually detected and utilized for training. In addition, we design an object instance reweighted loss to learn larger portion of each object instance to further improve the performance. The experimental results on two publicly available databases, VOC 2007 and 2012, demonstrate the efficacy of proposed approach.
\end{abstract}

\section{Introduction}

Object detection has always been one of the most essential technologies in computer vision field. Deep learning techniques introduced in recent years have significantly boosted state-of-the-art approaches for object detection \cite{girshick2015fast,liu2016ssd,redmon2016you,ren2015faster}. However, these approaches usually require large-scale manually annotated datasets \cite{russakovsky2015imagenet}. The high-cost of time-consuming accurate bounding box annotations, has impeded the wide deployment of CNN-based object detection technologies in real applications.

\begin{figure}[t]
\centering
\subfigure{
\includegraphics[scale = .31]{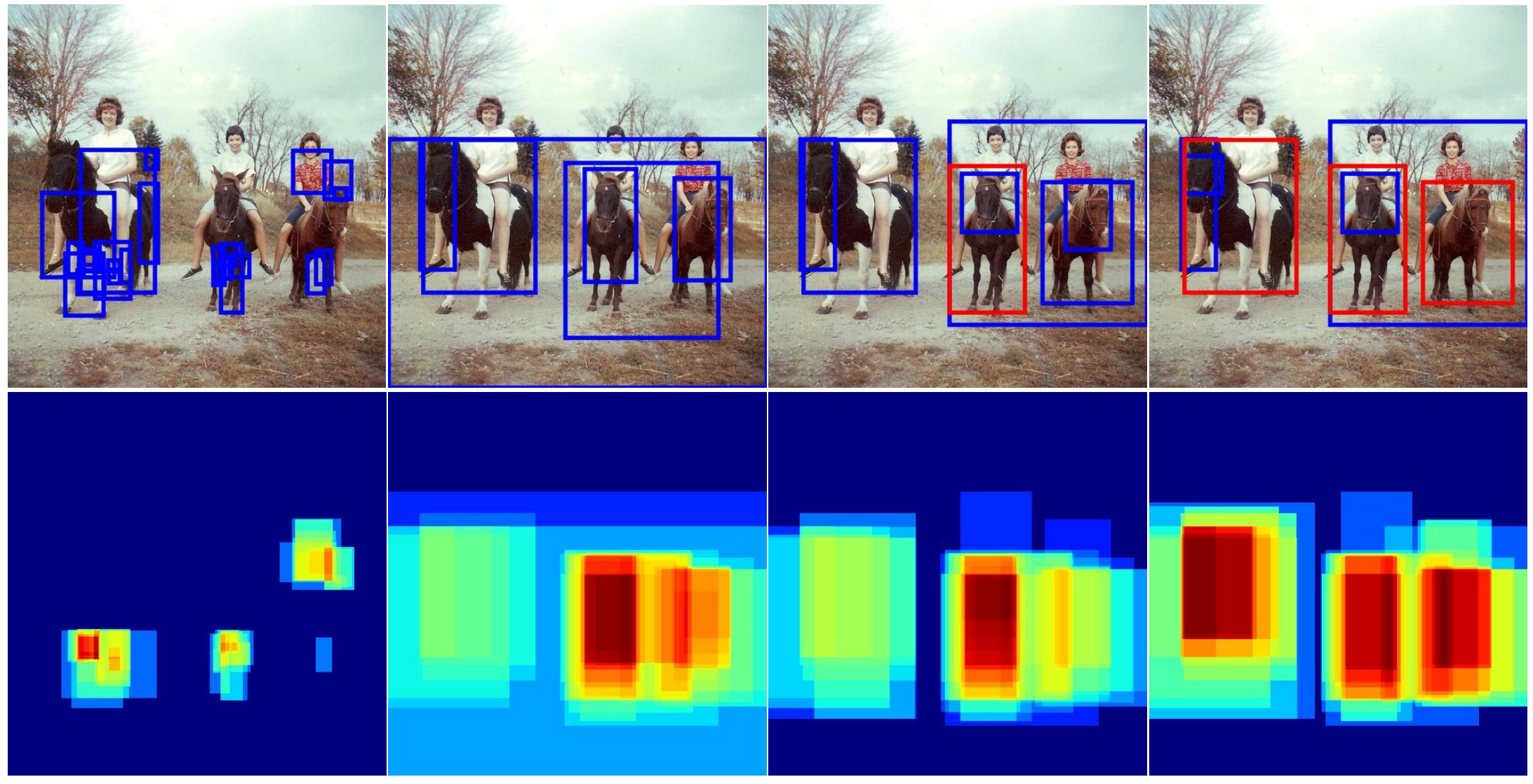}
}
\subfigure{
\includegraphics[scale = .31]{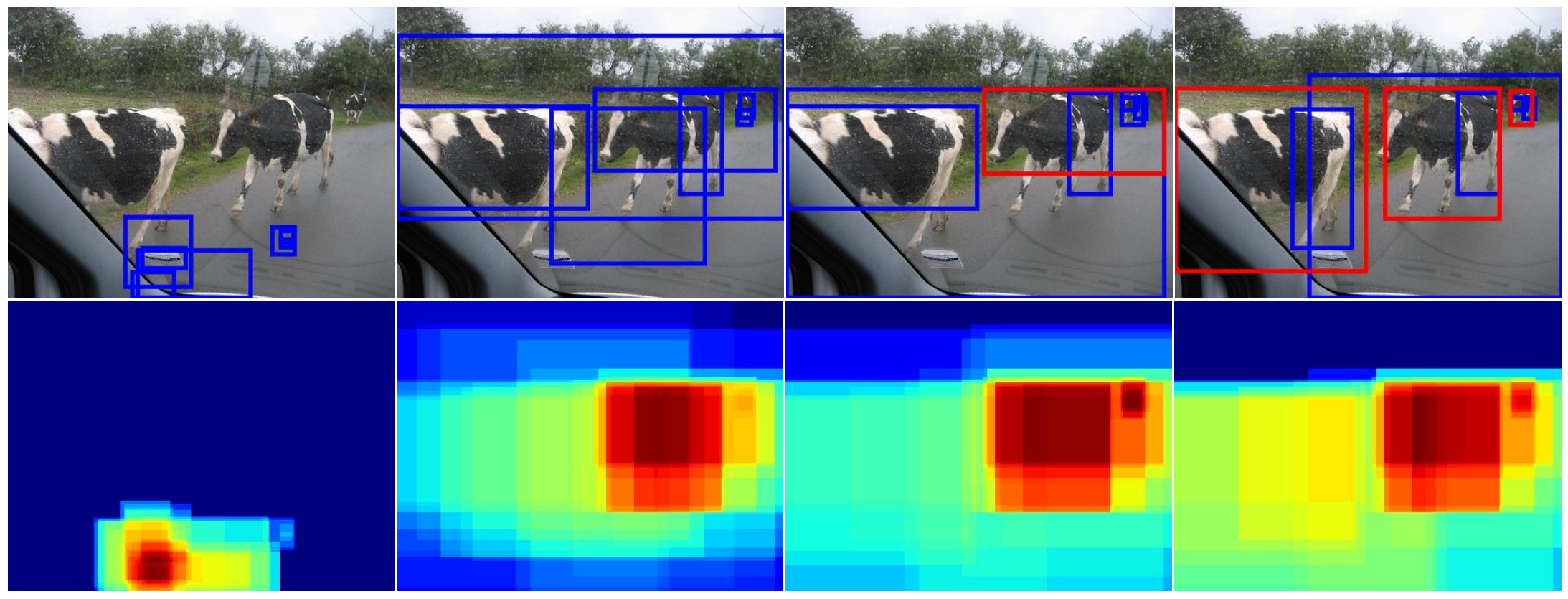}
}
\caption{The original images and corresponding objectness maps to show the evolution of object instance mining during learning process (from left to right). The first to fourth columns represent random initialization, epoch1, epoch3, and final epoch, respectively. Blue or red bounding boxes indicate the detected instances (top-scoring proposals after NMS) with detection scores $<0.5$ or $\geq0.5$.}
\label{fig-evolution}
\end{figure}

To address this limitation, weakly supervised object detection (WSOD) technology, which requires only image-level labels for training, has been introduced and explored \cite{bilen2016weakly,diba2017weakly,jie2017deep,oquab2015object,zhang2018w2f,tang2017multiple,zhang2018zigzag,shen2018generative,arun2019dissimilarity,pan2019low}. Although many approaches have been developed for WSOD and achieved promising results, the lack of object instance level annotations in images leads to huge performance gap between WSOD and fully supervised object detection (FSOD) methods.

Most previous approaches follow the framework of combining multiple instance learning (MIL) with CNN. This framework usually mines the most confident class-specific object proposals for learning CNN-based classifier, regardless of the number of object instances appearing in an image. For the images with multiple object instances from the same class, the object instances (fully annotated with bounding boxes in FSOD) with lower class-specific scores will be probably regarded as background regions. Many images in the challenging VOC datasets contain more than one object instance from the same class. For example, in VOC2007 \textit{trainval} set the number of image-level object labels and the annotated object instances are 7,913 and 15,662 respectively, which indicates that at least 7,749 instances are NOT selected during training. In this case, the selected object instances with relatively limited scale and appearance variations, may not be sufficient for training a CNN classifier with strong discriminative power. Moreover, the missing instances may be selected as negative samples during training, which may further degrades the discriminative capability of the CNN classifier. 

In this paper, an end-to-end object instance mining (OIM) framework is proposed to address the problem of multiple object instances in each image for WSOD. OIM is based on two fundamental assumptions: 1) the highest confidence proposal and its surrounding highly overlapped proposals
should probably belong to the same class; 2) the objects from the same class should have high appearance similarity. Formally, spatial and appearance graphs are constructed and utilized to mine all possible object instances present in an image and employ them for training. The spatial graph is designed to model the spatial relationship between the highest confidence proposal and its surrounding proposals, while the appearance graph aims at capturing all possible object instances having high appearance similarities with the most confident proposal. By integrating these two graphs into the iterative training process, an OIM approach that attempts to accurately mine all possible object instances in each image with only image-level supervision is proposed. With more object instances for training, a CNN classifier can have stronger discriminative power and generalization capabilities. The proposed OIM can further prevent the learning process from falling into local optima because more objects per-class with high similarity are employed for training. The original images and the corresponding objectness maps shown in Figure \ref{fig-evolution} illustrate that with the increasing number of iterations, multiple object instances belonging to the same class can be detected and are employed for training using the proposed approach.

Another observation from existing approaches is that the most confident region proposal is easy to concentrate on the locally distinct part of an object, especially for non-rigid objects such as human and animals. This may lead to the problem of detecting only small part of the object. To alleviate this problem, an object instance reweighted loss using the spatial graph is presented to help the network detect more accurate bounding box. This loss tends to make the network pay less attention on the local distinct parts and focus on learning the larger portion of each object. 

Our key contributions can be summarized as follows: 

\begin{itemize}
\item An object instance mining approach using spatial and appearance graphs is developed to mine all possible object instances with only image-level annotation, and it can significantly improve the discriminative capability of the trained CNN classifier. 
\item An object instance reweighted loss by adjusting the weight of loss function of different instances is proposed to learn more accurate CNN classifier. 
\end{itemize}

\begin{figure*}[t]
\centering
  \includegraphics[scale = .41]{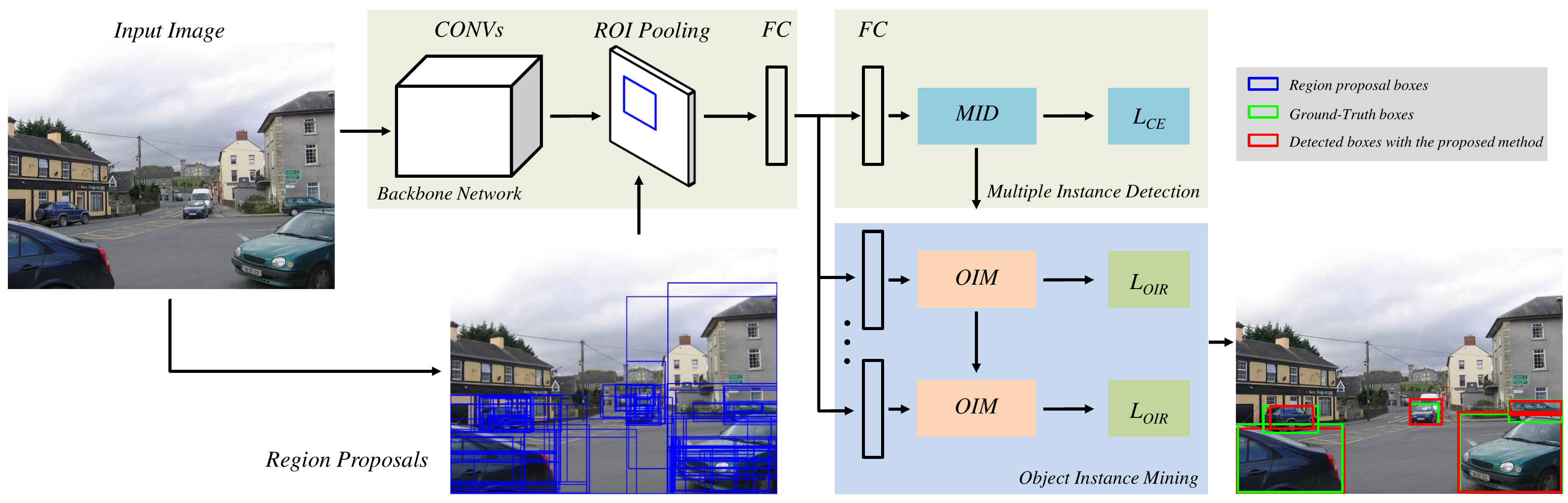}\\
\caption{Architecture of the proposed object instance mining framework. \textit{MID} represents multiple instance detector and \textit{OIM} indicates proposed object instance mining. $L_{\textit{CE}}$ is multi-class cross entropy loss and $L_{\textit{OIR}}$ is proposed instance reweighted loss.}
\label{fig-architecture}
\end{figure*}

\section{Related Work}
With only image-level annotations, most existing approaches implement weakly supervised object detection \cite{bilen2016weakly,tang2017multiple,jie2017deep,wan2019c} through multiple instance learning (MIL) framework \cite{dietterich1997solving}. The training images are firstly divided into bag of proposals (instances) containing positive target objects and negative backgrounds and CNN classifier is trained to classify the proposals into different categories. The most discriminative representation of instances is easy to be distinguished by such classifier that may make network trap into local optima. 

Recently, Bilen et al. \cite{bilen2016weakly} proposed a weakly supervised deep detection network (WSDDN) to perform object localization and classification simultaneously. Following this work, Tang \textit{et al.} \cite{tang2017multiple} introduced an online instance classifier refinement (OICR) strategy to learn larger portion of the objects. Such approach improves the performance of WSOD. However, it is also easy to trap into local optima since only the most discriminative instance is selected for refinement. 
Wan \textit{et al.} \cite{wan2018min} developed a min-entropy latent model to classify and locate the objects by minimizing the global and local entropies, which was proved to effectively boost the detection performance. Wan \textit{et al.} \cite{wan2019c} also attempted to address the local minima problem in MIL using continuation optimization method. In references \cite{wei2018ts2c,shen2019cyclic,li2019weakly}, the authors attempted to integrate segmentation task into weakly supervised object detection to obtain more accurate object bounding boxes. However, these methods require complex training framework with high training and test time complexity.

The authors in \cite{tang2018pcl} proposed to use proposal cluster to divide all proposals into different small bags and then classifier refinement was applied. This approach attempted to classify and refine all possible objects in each image. However, many proposals containing part of the object might be ignored using proposal cluster during the training. Gao et al. \cite{gao2018c} introduced a count-guided weakly supervised localization approach to detect per-class objects in each image. A simple count-based region selection algorithm was proposed and integrated into OICR to improve the performance of WSOD. However, the extra count annotations which needs a certain human labor are introduced and their method requires an alternative training process which can be time-consuming. In this paper, the count annotation is replaced by the proposed OIM algorithm without extra labor cost.

\section{Proposed Approach}

\subsection{Overall Framework}
The overall architecture of the proposed framework illustrated in Figure \ref{fig-architecture} mainly consists of two parts. The first part is a multiple instance detector (MID) which is similar to the structure presented in \cite{bilen2016weakly}. It performs region selection and classification simultaneously using a weighted MIL pooling. The second part is the proposed object instance mining and the proposed instance reweighted loss. During the training phase, we firstly adopt MID to classify the region proposals into different predicted classes. Then the detection outputs and proposal features are integrated to search all possible object instances from the same class in each image using spatial and appearance graphs. In addition, the instance reweighted loss is designed to learn larger portion of each object. As can be seen from the Figure \ref{fig-architecture}, the multiple object instances belonging to the same class can be accurately detected using the proposed method.
\subsection{Object Instance Mining}
\label{instance}
Previous methods \cite{tang2017multiple,gao2018c,zhang2018w2f,wei2018ts2c} often select the most confident proposal from each class as the positive sample to refine the multiple instance detector. The performance improvement can be limited using these methods, since only the top-scoring and surrounding proposals are selected for refinement. While in many conditions, there are multiple object instances belonging to the same class in an image. Those ignored object instances may be regarded as negative samples during the training that may degrade the performance of WSOD. Therefore, we propose an object instance mining (OIM) approach by building spatial graphs and appearance graphs to search all possible object instances in each image and integrate them into the training process.

Based on the assumption that the top-scoring and surrounding proposals with large overlaps (spatial similarity) should have the same predicted class, the spatial graphs can be built. We also assume that the objects from the same class should have similar appearance. Based on the similarities between the top-scoring proposal and the other proposals, the appearance graphs are built. Then we search all possible object instances in each image and employ them for training through these graphs. 

Given an input image $I$ with class label $c$, a set of region proposals $\mbox{P}=\{\bm{p}_1, ..., \bm{p}_N\}$ and their corresponding confidence scores $\mbox{X}=\{\bm{x}_1, ..., \bm{x}_N\}$, the core instance (proposal) $\bm{p}_{i_c}$ with the highest confidence score $\bm{x}_{i_c}$ can be selected. Here $i_c$ donates the index of this core instance (proposal). The core spatial graph can be defined by $G^s_{i_c}=(V^s_{i_c},E^s_{i_c})$, where each node in $V^s_{i_c}$ represents a selected proposal which has the overlap, i.e. spatial similarity, with the core instance larger than a threshold $T$. Each edge in $E^s_{i_c}$ represents such spatial similarity. All the nodes in spatial graph $G^s_{i_c}$ will be selected and labelled to the same class as $\bm{p}_{i_c}$. 

\begin{figure*}[t]
\centering
\subfigure[]{
  \includegraphics[scale = .31]{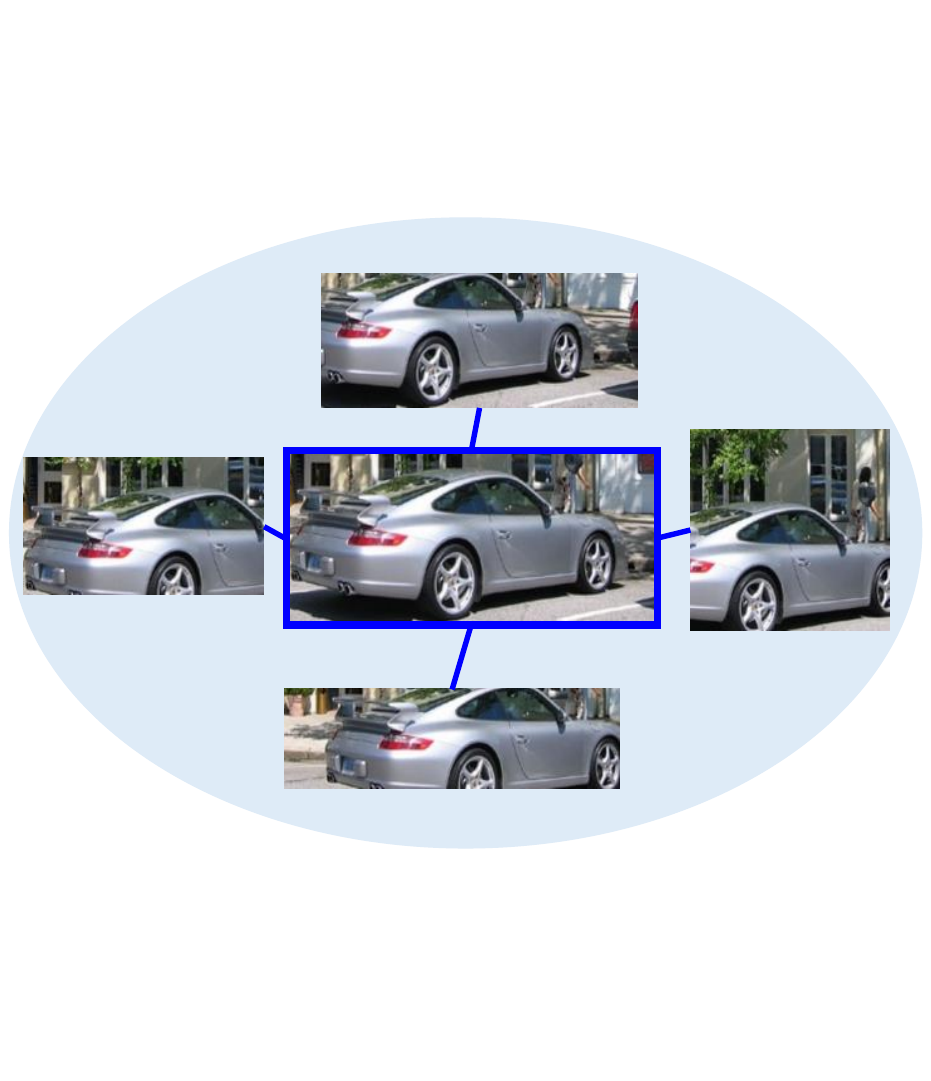}
}
\subfigure[]{
  \includegraphics[scale = .31]{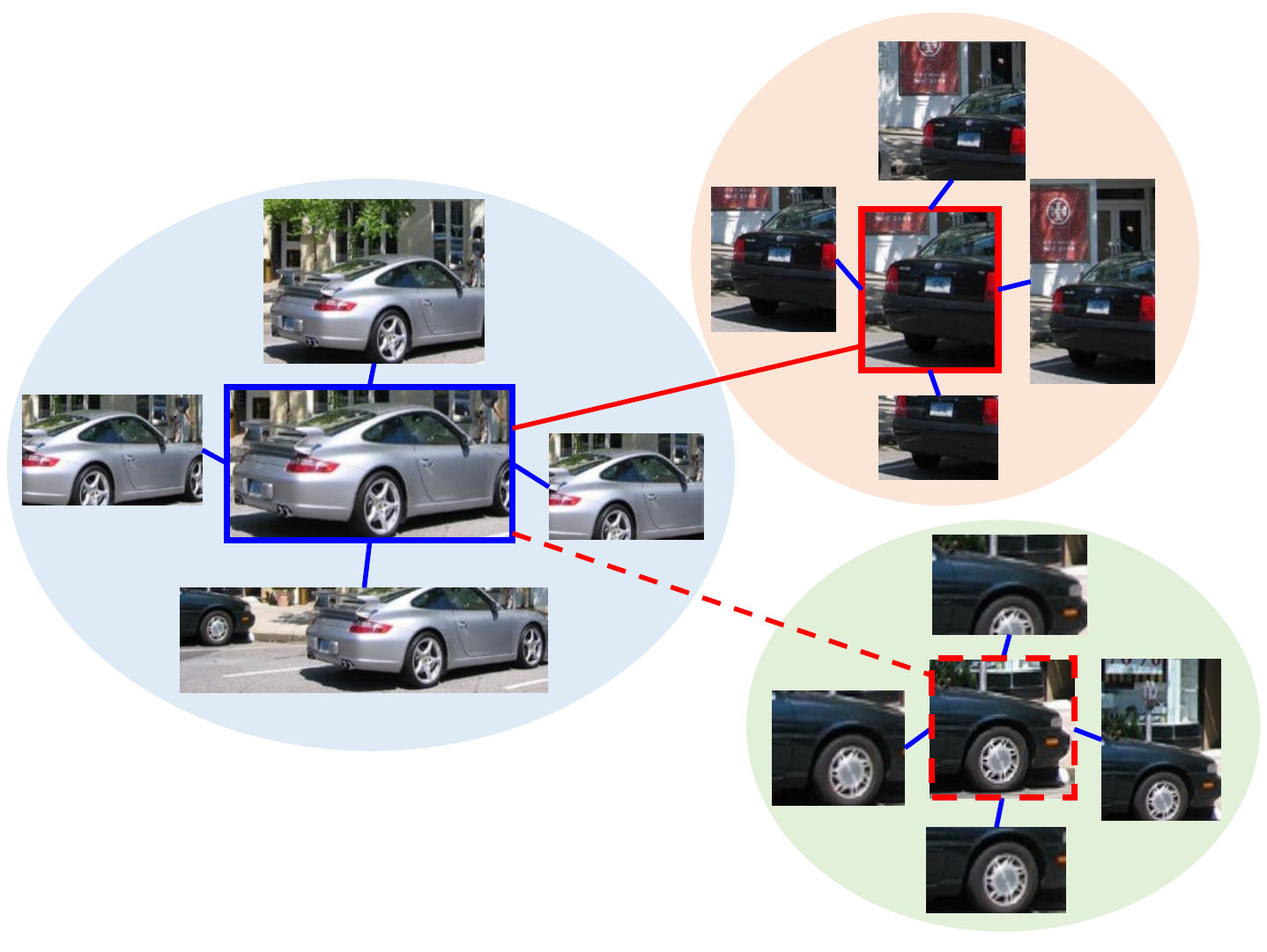}
}
\subfigure[]{
  \includegraphics[scale = .31]{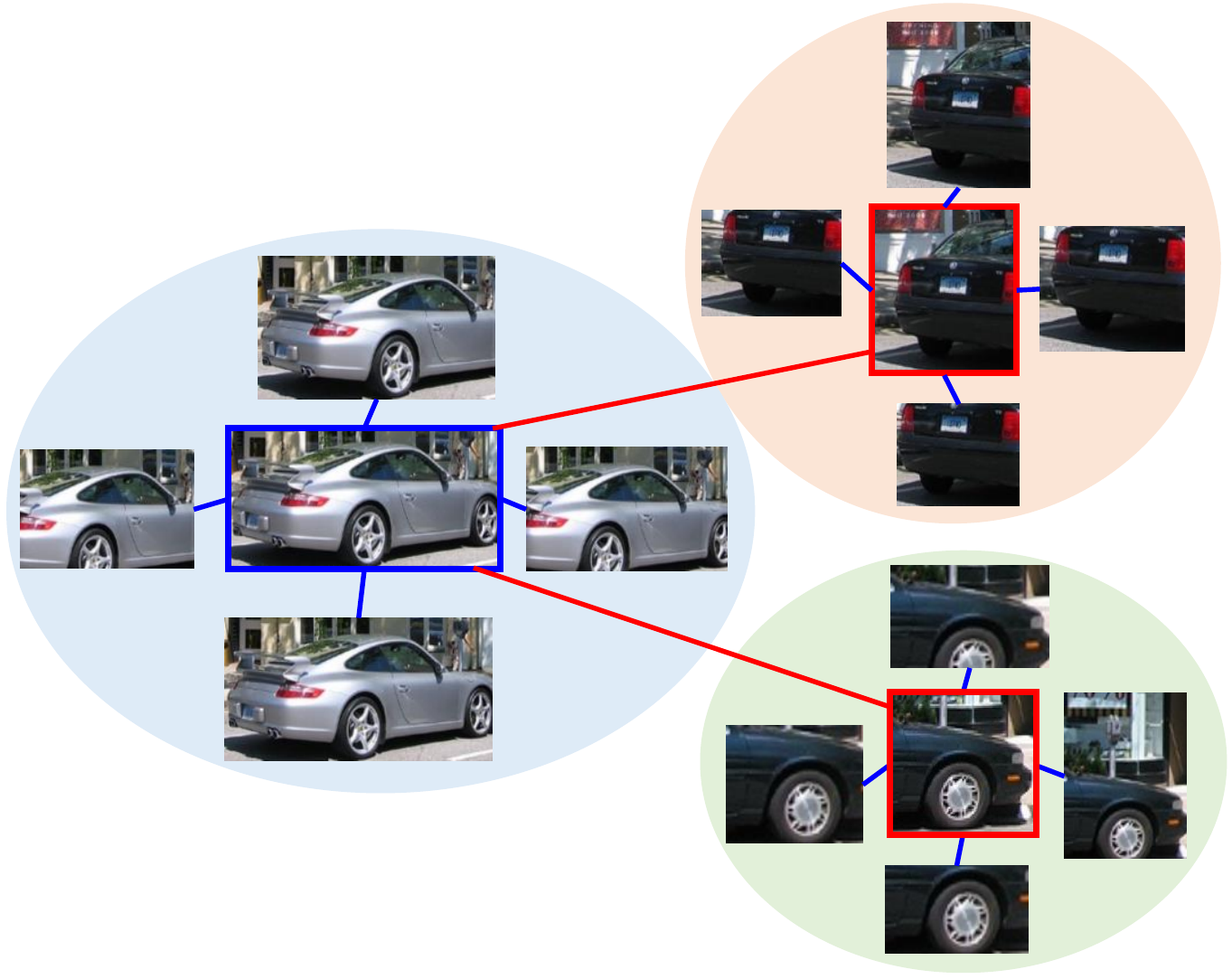}
}
\subfigure[]{
  \includegraphics[scale = .3]{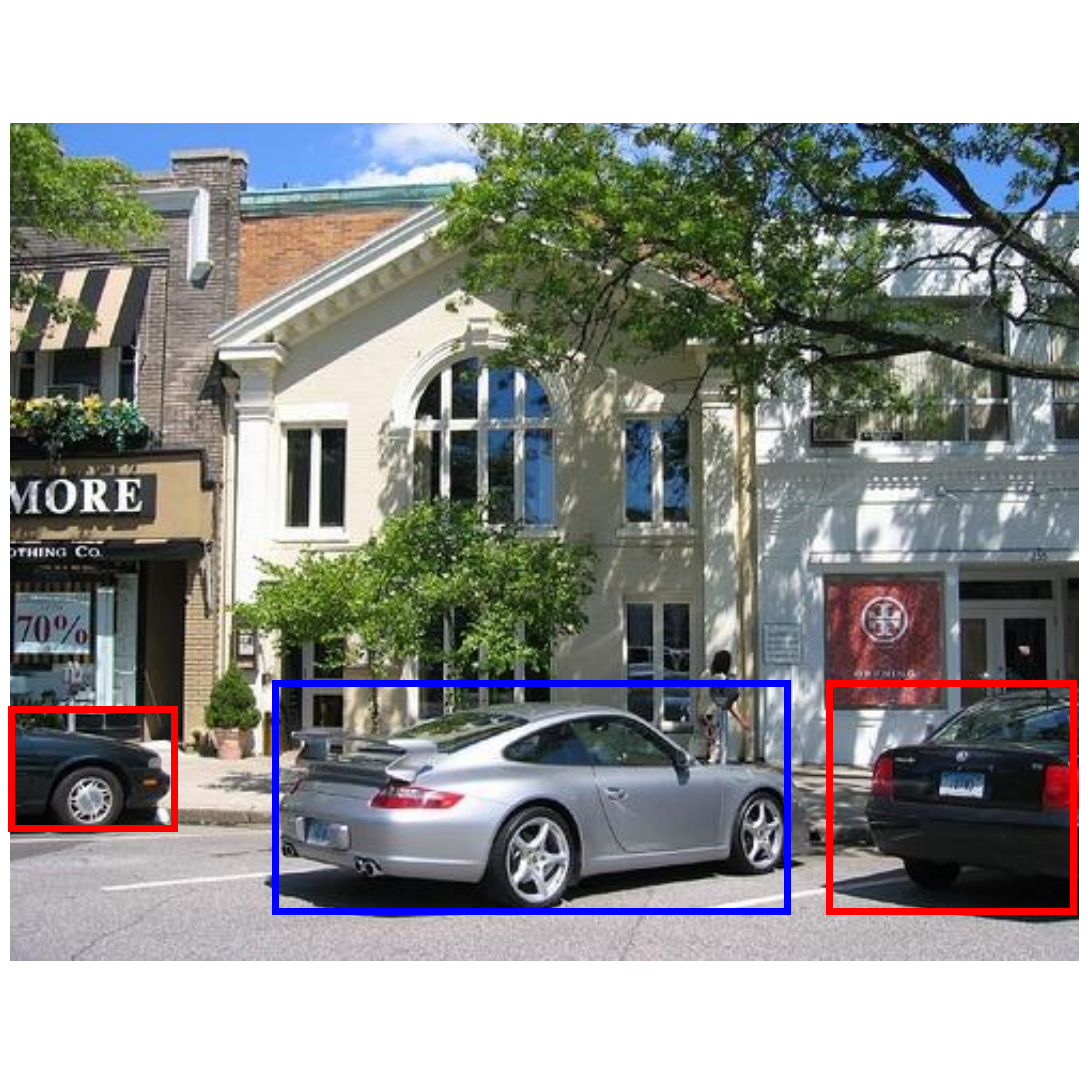}
}
\caption{Process to explore all possible object instances from the same class using OIM. (a)-(c) illustrate the spatial and appearance graphs of different epochs and (d) shows all detected instances. Blue bounding boxes represent the detected core instance with the highest confidence score. Red bounding boxes represent the other detected instances which have high appearance similarities with the core instance. Blue and red line represent spatial and appearance graph edge respectively. Red broken line in (b) means the appearance similarities is smaller than the threshold and thus the object instances are not employed in this stage.}
\label{fig-3}
\end{figure*}
We define feature vectors of each proposal as $\mbox{F}=\{\bm{f}_1, ..., \bm{f}_N\}$ and it can be generated from the fully connected layer. Each vector encodes a feature representation of a region proposal. Then the appearance graph is defined as $G^a =(V^a,E^a)$, where each node in $V^a$ is a selected proposal which has high appearance similarity with the core instance and each edge in $E^a$ represents the appearance similarity. This similarity can be calculated from the feature vectors of core instance and one of the other proposals (e.g. $\bm{p}_j$) using the Euclidean distance, denoted as follows,
\begin{equation}
{D}_{i_c,j} =  \left\|\bm{f}_{i_c} - \bm{f}_{j} \right\|_2.
\label{eq1}
\end{equation}
Only if the proposal $\bm{p}_j$ meets the condition that $D_{i_c,j}<$ $\alpha D_{avg}$ and $\bm{p}_j$ has no overlap with all the proposals previously selected, such proposal will be added into the nodes in $G^a$. $D_{avg}$ represents the average inter-class similarity of the core spatial graph $G^s_{i_c}$ using average distance of all the nodes in $G^s_{i_c}$ and it can be defined as follows,
\begin{equation}
\begin{aligned}
{D}_{\text{avg}}=\dfrac{1}{M}\sum_{k} {D}_{i_c,k},
{\kern 1pt}{\kern 1pt}{\kern 1pt}{\kern 1pt}{\kern 1pt}\\ 
{\rm{s}}{\rm{.t}}{\rm{.}}{\kern 1pt}{\kern 1pt}{\kern 1pt}{\kern 1pt}IoU(\bm{p}_{i_c},\bm{p}_k)>T.
\end{aligned}
\end{equation}
where $\bm{p}_k$ represents the node meet the constraints above and $M$ indicates the number of these nodes in $G^s_{i_c}$. $\alpha$ is a hyper parameter which is determined by experiments.
\begin{algorithm}[t]
\label{algorithm}
\footnotesize
\SetAlgoNoLine
\caption{Object Instance Mining}
\KwIn{Image $I$, region proposals $\mbox{P}=\{\bm{p}_1, ..., \bm{p}_N\}$, image label $\mbox{Y}=\{y_1,y_2,...y_c\}$}
\KwOut{All the nodes $V^a$ in appearance graph}
Feed Image $I$ and its proposals into the network to produce feature vectors $\mbox{F}=\{\bm{f}_1, ..., \bm{f}_N\}$ \\
\For{$c$ \rm{in} $C$, $C$ \rm{is the list of training data class}}{
    \If{$y_c==1$}{
        $V^s \gets \emptyset$, $V^a \gets \emptyset$,$D \gets 0$, $D_{avg} \gets 0$, $M\gets 0$, $flag \gets 0$\\
        Choose the top-scoring proposal $i_{c}$\\
        $V^s_{i_c} \gets \bm{p}_{i_c}$, $V^a \gets \bm{p}_{i_c}$\\
        \For{$j=$ \rm{1 to }$N$}{
			 Compute the appearance similarity $D_{i_c,j}$ using Eq. \ref{eq1} \\
            Compute $IoU$($\bm{p}_{i_c}$,$\bm{p}_j)$ \\
            \If{$IoU$($\bm{p}_{i_c}$,$\bm{p}_j)>T$}{
			     $V^s_{i_c} \gets \bm{p}_{j}$\\$D \gets  D_{i_c,j}+D$,$M \gets M + 1$\\
                }
            }
            $D_{avg} \gets \dfrac{D}{M}$\\
        Sort (ascend) $\mbox{P}$ based on $D_{i_c,j}$\\
        \For{$j=$ \rm{1 to }$N$}{
            \If{$D_{i_c,j}<$ $\alpha D_{avg}$}{
                \If{$\exists$ $\bm{p}_k$ $\in$ $V^a$, \rm $IoU$($\bm{p}_{k}$,$\bm{p}_j)>$ 0}{
                        $flag \gets 1$\\
                }
                \If{flag $==$ \rm{0}}{$V^a \gets \bm{p}_{j}$, $V^s_j \gets \bm{p}_{j}$ 
}
            }
        }
    }
}
\end{algorithm}

The proposed object instance mining (OIM) approach using spatial and appearance graphs is summarized in Algorithm \ref{algorithm}. We also build spatial graph $G^s$ for each node in appearance graph $G^a$ and then all these nodes will be included for training. If no proposal has high similarity with the core instance, only the core instance and surrounding proposals, i.e. spatial graph $G^s_{i_c}$ will be employed. In such a way, more instances from the same class with similar appearance and different poses will be employed for training. It results in that not only more object instances can be detected but also more accurate detected boxes can be learned. 

Figure \ref{fig-3} illustrates the process to detect all possible object instances from to the same class using spatial and appearance graphs. Figure \ref{fig-3} (a) is the core spatial graph and figure \ref{fig-3} (b)-(c) describe the spatial and appearance graphs in different epochs. With the increased number of iterations, more instances can be detected using the proposed OIM.

\subsection{Instance Reweighted Loss}
In addition to exploring all possible object instances in each image, we also design an object instance reweighted loss to learn more accurate detected boxes. During the iterative learning process, the CNN-based classifier is easy to learn the most distinct part of each object instance instead of the whole body, especially for the non-rigid one. We propose to assign different proposal weights to individual proposals to balance the weight of the top-scoring proposal and surrounding less discriminative ones. Thus the larger portion of each instance is expected to be detected. 

Given an image with label $\mbox{Y}$ and predicted label $\mbox{Y}_j = [y_{0,j},y_{1,j},...,y_{C,j}]^T \in \mathbb{R}^{(C+1)\times1}$ for the $j$-th proposal in a spatial graph $G^s$, where $y_{c,j}=1$ or $0$ indicates the proposal belonging to class $c$ or not, and $c=0$ is index of background class. The loss in Eq.~\ref{oicr-refinment-loss} is similar to the loss in \cite{tang2017multiple}, where $w_j$ is the loss weight of $j$-th proposal.
$x_{c,j}^s$ with class label $c$ in $G^s$, are the proposals used for training and $x_{c,i_c}^s$ is center (core) proposal with the highest score.
\begin{equation}
  \label{oicr-refinment-loss}
  \mathcal{L} = - \frac{1}{|\mbox{P}|} \sum_{j=1}^{|\mbox{P}|} \sum_{c=1}^{C+1} w_j y_{c,j} \log x_{c,j}^s.
\end{equation}
\begin{figure*}[t!]
\centering
  \includegraphics[scale = 0.88]{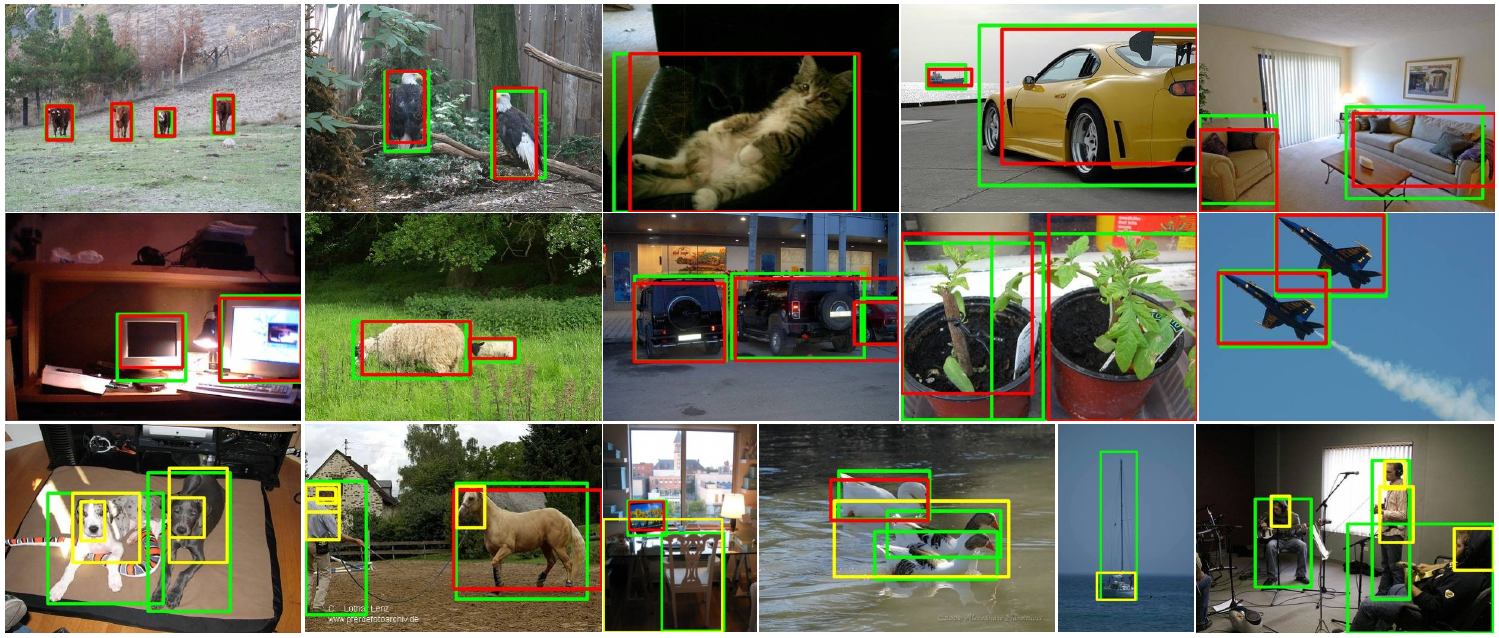} \\
\caption{Detections examples on VOC2007 \textit{test} set. The green bounding boxes represent the ground-truth. The successful detections (IoU $\geq$ 0.5) are marked with red bounding boxes, and the failed ones are marked with yellow color. We show all detections with scores $\geq$ 0.5 and NMS is performed to remove duplicate detections.}
\label{figure-voc2007test}
\end{figure*}
It can be seen from Eq.~\ref{oicr-refinment-loss} that proposals in each spatial graph contribute equally. Thus, the non-center proposals with relative low scores in each spatial graph are difficult to be learned during training.
To address this problem, an instance reweighted loss function is designed as follows,
\begin{equation}
  \label{loss_max}
  \mathcal{L} = - \frac{1}{|\mbox{P}|} \sum_{j=1}^{|\mbox{P}|} \sum_{c=1}^{C+1} w_j y_{c,j} (1 + z_{j}^s) \log x_{c,j}^s,
\end{equation}

\noindent where $z_{j}^s$ is introduced to balance the proposal weights in spatial graph $G^s$ as defined in Eq.~\ref{maximum}. $\beta$ is hyper-parameter.
\begin{equation}
  \label{maximum}
  z_{j}^s =
  \left\{
    \begin{array}{lr}
      \beta, & j \neq i_c \\
      \\
      \beta-1, & j = i_c \\
    \end{array}
  \right.
\end{equation}
To guide the network to pay more attention on learning the less discriminative regions of the object instance in each graph $G^s$, we balance the weight of the surrounding less discriminative proposals with the center proposal using Eq.~\ref{loss_max} and Eq.~\ref{maximum}. As a result, gradients of surrounding proposals are scaled up to $(1 + \beta)$ of its original value, while gradient of the center proposal is scaled to $\beta$ of its original value during back-propagation. 
Similar to the implementation in \cite{gao2018c}, we also use the standard multi-class cross entropy loss for the multi-label classification and it is combined with the proposed instance reweighted loss for training.

\section{Experiments}

\subsection{Datasets and Evaluation Metrics}
Following the previous state-of-the-art methods on WSOD, we also evaluate our approach two datasets, PASCAL VOC2007\cite{everingham2010pascal} and VOC2012\cite{everingham2015pascal}, which both contain 20 object categories. For VOC2007, we train the model on the \textit{trainval} set (5,011 images) and evaluate the performance on the \textit{test} set (4,952 images). For VOC2012, the \textit{trainval} set (11,540 images) and the \textit{test} set (10,991 images) are used for training and evaluation respectively. Additionally, we train our model on the VOC2012 \textit{train} set (5,717 images) and proceed evaluation on the \textit{val} set (5,823 images) to further validate the effectiveness of proposed approach. Following previous work, we use mean average precision (mAP) to evaluate the performance of proposed approach. Correct localization (CorLoc) is applied to evaluate the localization accuracy.

\subsection{Implementation Details}
To make a fair comparison, VGG16 model pre-trained on the ImageNet dataset \cite{russakovsky2015imagenet} is adopted as the backbone network to finetune the CNN classifier. The object proposals are generated using Selective Search\cite{uijlings2013selective}. The batch size is set to 2, and the learning rates are set to 0.001 and 0.0001 for the first 40K and the following 50K iterations respectively. During training and test, we take five image scales \{480, 576, 688, 864, 1200\} along with random horizontal flipping for data augmentation. Following \cite{tang2017multiple}, the threshold $T$ is set to 0.5. With the increased number of iterations, the network has more stable learning ability, we dynamically set the hyper parameters $\alpha$ as $\alpha_1=5$ for the first 70K and $\alpha_2=2$ for the following 20K iterations. $\beta$ are empirically set to 0.2 in our experiments. We also analyze the influence of these parameters in the ablation experiments section. 100 top-scoring region proposals are kept and Non-Maximun Suppression with IoU of 0.3 per class is performed to calculate mAP and CorLoc.

\begin{table*}[t!]
\caption{Comparison with the state-of-the-arts in terms of mAP (\%) on the VOC2007 \textit{test} set.}
\label{voc2007test}
\scriptsize
\begin{center}
\resizebox{\textwidth}{!}{
\begin{tabular}{|l|cccccccccccccccccccc|l|c|l|}
\hline
Method & aero & bike & bird & boat & bottle & bus & car & cat & chair & cow & table & dog & horse & mbike & person & plant & sheep & sofa & train & tv & mAP \\ \hline
OICR & 58.0 & 62.4 & 31.1 & 19.4 & 13.0 & 65.1 & 62.2 & 28.4 & 24.8 & 44.7 & 30.6 & 25.3 & 37.8 & 65.5 & 15.7 & 24.1 & 41.7 & 46.9 & 64.3 & 62.6 & 41.2 \\
PCL & 54.4& 69.0& 39.3 & 19.2& 15.7& 62.9& 64.4& 30.0& 25.1& 52.5 &44.4 & 19.6 & 39.3 & 67.7 & 17.8& 22.9 & 46.6& {57.5} & 58.6 & 63.0 & 43.5 \\ 
TS$^2$C & 59.3 & 57.5 & 43.7 & 27.3 & 13.5 & 63.9 & 61.7 & 59.9 & 24.1 & 46.9 & 36.7 & 45.6 & 39.9 & 62.6 & 10.3 & 23.6 & 41.7 & 52.4 & 58.7 & 56.6 & 44.3 \\
C-WSL*  & 62.9 & 64.8 & 39.8 & 28.1 & 16.4 & 69.5 & 68.2 & 47.0 & \textbf{27.9} & 55.8 & 43.7 & 31.2 & 43.8 & 65.0 & 10.9 & 26.1 & 52.7 & 55.3 & 60.2 & 66.6 & 46.8 \\
MELM  & 55.6 & 66.9 & 34.2 & {29.1} & 16.4 & 68.8 & 68.1 & 43.0 & 25.0 & {65.6} & 45.3 & {53.2} & 49.6 & 68.6 & 2.0 & 25.4 & 52.5 & 56.8 & 62.1 & 57.1 & 47.3 \\
OICR+W-RPN  & - & - & - & - & - & - & - & - & - & - & - & - & - & - & - & - & - & - & - & -& 46.9 \\
SDCN  & 59.8 & 67.1 & {32.0} & \textbf{34.7} & 22.8 & 67.1 & 63.8 & 67.9 & 22.5 & 48.9 & {47.8} & 60.5 & {51.7} & 65.2 & {11.8} & 20.6 & 42.1 & 54.7 & 60.8 & 64.3 & 48.3 \\
WS-JDS  & 52.0 & 64.5 & {45.5} & 26.7 & {27.9} & 60.5 & 47.8 & 59.7 & 13.0 & 50.4 & {46.4} & 56.3 & {49.6} & 60.7 & {25.4} & 28.2 & 50.0 & 51.4 & 66.5 & 29.7 & 45.6 \\
C-MIL  & 62.5 & 58.4 & {49.5} & 32.1 & 19.8 & 70.5 & 66.1 & 63.4 & 20.0 & 60.5 & {52.9} & 53.5 & {57.4} & 68.9 & {8.4} & 24.6 & 51.8 & 58.7 & 66.7 & 63.5 & \textbf{50.5} \\
\textbf{OIM} & {62.2} & {67.2} & 48.0 & 29.6 & 23.5 & 68.7 & 69.3 & 64.3 & 22.8 & 59.6 & 39.6 & 30.7 & 42.7 & 69.8 & 3.1 & 23.3 & \textbf{57.9} & 55.4 & 63.4 & 63.5 & 48.2 \\
\textbf{OIM+IR} & 55.6 & 67.0 & 45.8 & 27.9 & {21.1} & 69.0 & {68.3} & {70.5} & {21.3} & 60.2 & 40.3 & 54.5 & 56.5 & {70.1} & 12.5 & 25.0 & 52.9 & 55.2 & {65.0} & 63.7 & \textbf{50.1} \\
\hline
C-WSL*+FRCNN  & 62.9 & 68.3 & 52.9 & 25.8 & 16.5 & 71.1 & 69.5 & 48.2 & 26.0 & 58.6 & 44.5 & 28.2 & 49.6 & 66.4 & 10.2 & 26.4 & 55.3 & 59.9 & 61.6 & 62.2 & 48.2 \\
SDCN+FRCNN  & 61.1 & 70.6 & {40.2} & {32.8} & 23.9 & 63.4 & 68.9 & 68.2 & 18.3 & 60.2 & {53.5} & 63.6 & {53.6} & 66.1 & {14.6} & 21.8 & 50.5 & 56.7 & 62.4 & \textbf{67.9} & 51.0 \\
WS-JDS+FRCNN  & 64.8 & 70.7 & {51.5} & 25.1 & \textbf{29.0} & {74.1} & 69.7 & 69.6 & 12.7 & \textbf{69.5} & {43.9} & 54.9 & {39.3} & \textbf{71.3} & \textbf{32.6} & \textbf{29.8} & 57.0 & 61.0 & 66.6 & 57.4 & 52.5 \\
Pred Net (FRCNN) & \textbf{66.7} & 69.5 & {52.8} & 31.4 & 24.7 & \textbf{74.5} & \textbf{74.1} & 67.3 & 14.6 & 53.0 & {46.1} & 52.9 & \textbf{69.9} & 70.8 & {18.5} & 28.4 & 54.6 & 60.7 & 67.1 & 60.4 & \textbf{52.9} \\
C-MIL+FRCNN & 61.8 & 60.9 & \textbf{56.2} & 28.9 & 18.9 & 68.2 & 69.6 & 71.4 & 18.5 & 64.3 & \textbf{57.2} & \textbf{66.9} & {65.9} & 65.7 & {13.8} & 22.9 & 54.1 & \textbf{61.9} & 68.2 & 66.1 & \textbf{53.1} \\
\textbf{OIM+IR+FRCNN} & 53.4 & \textbf{72.0} & 51.4 & 26.0 & {27.7} & 69.8 & {69.7} & \textbf{74.8} & {21.4} & 67.1 & 45.7 & 63.7 & 63.7 & {67.4} & 10.9 & 25.3 & 53.5 & 60.4 & \textbf{70.8} & 58.1 & \textbf{52.6} \\
\hline
\end{tabular}}
\end{center}
\end{table*}

\subsection{Comparison with State-of-the-arts}
State-of-the-art WSOD methods are used for comparison to validate the effectiveness of the proposed approach. Table \ref{voc2007test} shows performance comparison in terms of mAP on VOC2007 \textit{test} set. By only using OIM, better or similar results can be achieved compared with previous SOTA methods such as MELM, SDCN, $etc$. We attribute this improvement to the OIM, which increases the representation capability of the trained CNN by searching more objects from the same class and employing them into training. As the detected bounding boxes and objectness maps shown in Figure \ref{fig-evolution}, the confidence scores of less discriminative objects are gradually improved and more objects from the same class can be detected during the training. It further proves that integrating the less discriminative objects into training improves the performance for WSOD. Further performance improvement can be achieved using the proposed instance reweighted loss. The proposed approach achieves a mAP of 50.1\%, which outperforms the PCL,C-WSL$^*$, SDCN, WS-JDS methods, $etc$, and the performance is similar to the result of C-MIL.
We further used the learned objects as pseudo ground-truth to train a Fast-RCNN-based detector, our approach also achieve better or similar performance as compared with previous state-of-the-art methods.

\begin{figure}[t!]
\centering
  \includegraphics[scale = 0.25]{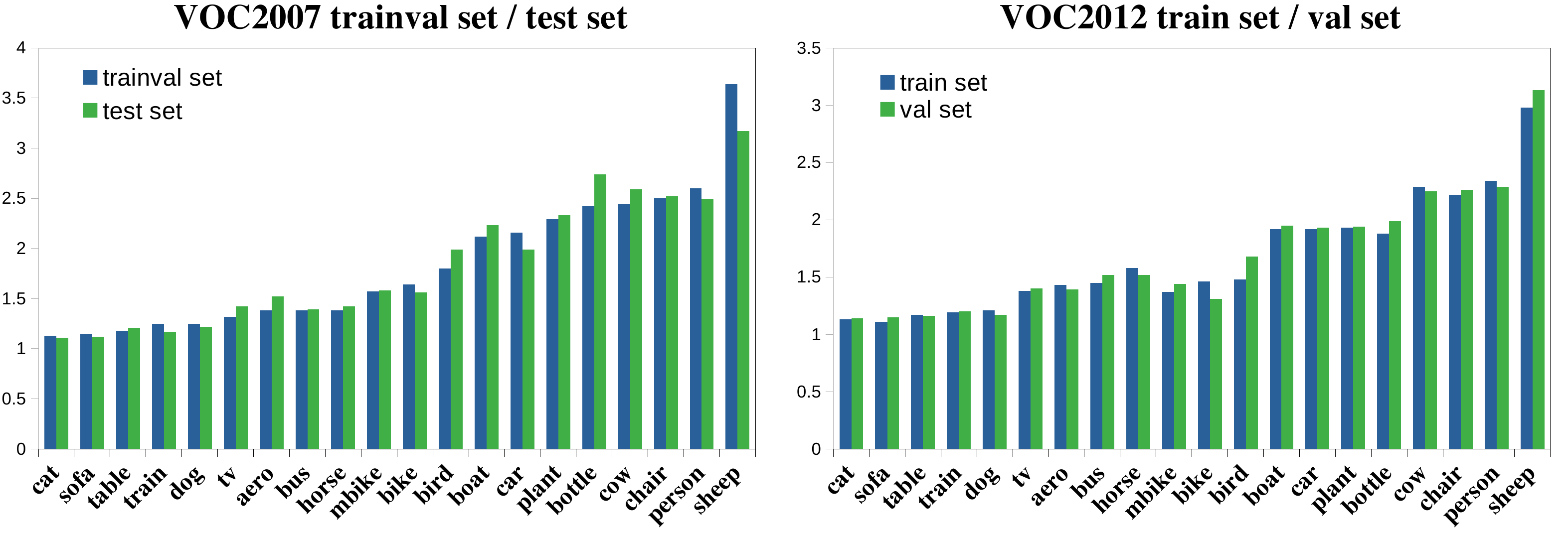} \\
\caption{Objects number of each class divided by the number of images which the corresponding class occurs on VOC2007 and VOC2012.}
\label{figure-chart}
\end{figure}

In particular, by only using the proposed OIM strategy, our approach outperforms C-WSL method by 1.4 \% without introducing extra per-class count supervision. 
Our work attempts to include all possible object instances from each class for training since many images contain more than one per-class object instance. Figure \ref{figure-chart} illustrates that most classes in two datasets have more than one object instance in an image. Specifically, almost half of categories contain more than two object instances in an image. Especially for class ``sheep'', which the average number of sheep appearing in an image is larger than 3, our OIM method (57.9 \% mAP) performs better than all the other methods. In addition, for most non-rigid objects (``cat'', ``dog'', ``horse'', ``person', $etc.$), as can be seen from Table \ref{voc2007test}, by applying instance reweighted loss more accurate object instance can be detected.

\begin{table}[t!]
\caption{Comparison with the state-of-the-arts in terms of CorLoc (\%) on the VOC2007 \textit{trainval} set.}
\label{voc2007trainval}
\scriptsize
\begin{center}
\resizebox{.95\columnwidth}{26mm}{
\begin{tabular}{|p{4.3cm}<{\centering}|p{2.3cm}<{\centering}|}
\hline
Method & Localization (CorLoc)   \\
\hline
OICR\cite{tang2017multiple} &  60.6   \\
PCL\cite{tang2018pcl} &    62.7   \\
C-WSL* \cite{gao2018c} & 63.5 \\
MELM \cite{wan2018min} & 61.4 \\
WS-JDS \cite{shen2019cyclic} & 64.5 \\ 
C-MIL \cite{wan2019c} & 65.0 \\
OICR+W-RPN \cite{singh2019you}  & 66.5 \\
SDCN \cite{li2019weakly} & 66.8 \\ 
\textbf{OIM+IR} & \textbf{67.2} \\
\hline
C-WSL*+FRCNN \cite{gao2018c} & 66.1\\
WS-JDS+FRCNN \cite{shen2019cyclic} & 68.6 \\ 
SDCN+FRCNN \cite{li2019weakly} & \textbf{68.8} \\
Pred Net (FRCNN) (Arun et al. 2019) & \textbf{70.9} \\
\textbf{OIM+IR+FRCNN} & \textbf{68.8}  \\
\hline
\end{tabular}}
\end{center}
\end{table}
%

\begin{table}[t!]
\caption{Comparison with the state-of-the-arts in terms of mAP (\%) on the VOC2012 \textit{test} set.}
\label{voc2012test}
\scriptsize
\begin{center}
\resizebox{.95\columnwidth}{25mm}{
\begin{tabular}{|p{4.7cm}<{\centering}|p{1cm}<{\centering}|p{1cm}<{\centering}|}
\hline
Method &Dataset  & mAP   \\
\hline
MELM \cite{wan2018min} & train/val &  40.2 \\
C-WSL \cite{gao2018c} &train/val & 43.0\\
\textbf{OIM+IR} &train/val & \textbf{44.4}\\
\hline
OICR \cite{tang2017multiple}&trainval/test &  37.9  \\
PCL \cite{tang2018pcl}& trainval/test &  40.6  \\
MELM \cite{wan2018min}& trainval/test &  42.4 \\
WS-JDS \cite{shen2019cyclic}& trainval/test &  39.1 \\
OICR+W-RPN \cite{singh2019you}& trainval/test  & 43.2 \\
SDCN \cite{li2019weakly}& trainval/test &  43.5 \\
\textbf{OIM+IR} & trainval/test & \textbf{45.3}\\
\hline
Pred Net (FRCNN) (Arun et al. 2019)& trainval/test & 48.4 \\
WS-JDS + FRCNN \cite{shen2019cyclic}& trainval/test & 46.1 \\
C-MIL + FRCNN \cite{wan2019c} & trainval/test & \textbf{46.7} \\
SDCN + FRCNN \cite{li2019weakly}& trainval/test & \textbf{46.7} \\
\textbf{OIM+IR + FRCNN} &trainval/test & \textbf{46.4}\\
\hline
\end{tabular}}
\end{center}
\end{table}

\begin{table}[t!]
\caption{Comparison with the state-of-the-arts in terms of CorLoc (\%) on the VOC2012 \textit{trainval} set.}
\label{voc2012trainval}
\scriptsize
\begin{center}
\resizebox{.95\columnwidth}{23mm}{
\begin{tabular}{|p{4.5cm}<{\centering}|p{2.3cm}<{\centering}|}
\hline
Method & Localization (CorLoc)   \\
\hline
OICR\cite{tang2017multiple} &    62.1   \\
PCL\cite{tang2018pcl} &    63.2   \\
WS-JDS \cite{shen2019cyclic} &  63.5 \\
OICR+W-RPN \cite{singh2019you} & \textbf{67.5} \\
SDCN \cite{li2019weakly} & \textbf{67.9} \\
\textbf{OIM+IR} & \textbf{67.1}\\
\hline
Pred Net (FRCNN)  (Arun et al. 2019) & 69.5 \\
WS-JDS + FRCNN \cite{shen2019cyclic} & 69.5 \\
C-MIL + FRCNN \cite{wan2019c} & {67.4} \\
SDCN + FRCNN \cite{li2019weakly} & {69.5} \\
\textbf{OIM+IR + FRCNN} & \textbf{69.5} \\
\hline
\end{tabular}}
\end{center}
\end{table}

CorLoc is also used as the evaluation metric to ascertain the performance of proposed method. Table \ref{voc2007trainval} shows performance comparison in terms of CorLoc on the VOC2007 \textit{trainval} set. Our result outperforms all existing state-of-the-art methods when Fast-RCNN detector is not used. The proposed OIM framework iteratively explores more object instances and larger portion of the instances from the same class with similar appearance and different poses for training, which makes more accurate detected boxes can be learned. Therefore, the proposed approach not only brings the mAP improvements but also makes the detected boxes more accurate which results in better CorLoc. 

The proposed approach is also evaluated on VOC2012 dataset. Since some approaches \cite{gao2018c} only use validation set of VOC2012 for evaluation, we use both \textit{test} and \textit{val} set to evaluate the proposed approach. In Table \ref{voc2012test}, the detection results in terms of mAP on \textit{test} and \textit{val} set are provided respectively. Table \ref{voc2012trainval} lists the CorLoc results on VOC2012 \textit{trainval} set. The experimental results in Tables \ref{voc2012test} and \ref{voc2012trainval} validate the effectiveness of the proposed approach.

Figure \ref{figure-voc2007test} visualizes the detection results on the VOC2007 \textit{test} set. The successful (IoU $\geq$ 0.5) and failed (IoU $<$ 0.5) detections are marked with red and yellow bounding boxes respectively. The green bounding boxes are the ground-truths. The first two rows indicate our approach can detect tight boxes even multiple objects from the same class co-occur in an image, e.g. ``cow'', ``sheep''. The last row shows some failed cases, which are often attribute to localizing the most discriminative parts of non-rigid objects, grouping multiple objects, and background clutter, e.g.  ``human''.

\begin{table}[t!]
  \caption{Detection performance (mAP\%) on the VOC2007 for using different values of parameter $\alpha$ and parameter $\beta$.}
  \label{tab1}
 \scriptsize
  \begin{center}
\begin{minipage}{0.5\linewidth}
  \centerline{\begin{tabular}{|p{1cm}<{\centering}|p{1cm}<{\centering}|p{1cm}<{\centering}|p{1cm}<{\centering}|p{1cm}<{\centering}|}
  \hline
  $\alpha_1$ &1&2&5&10\\
  \hline
  $\alpha_2$ &1&2&2&5\\
  \hline
  OIM & 42.9 & {48.1} & \textbf{48.2} & {46.8} \\
  \hline
  OIM+IR & 43.4 & {49.3} & \textbf{50.1} & {48.4} \\
  \hline
  \end{tabular}}
\end{minipage}
\begin{minipage}{0.5\linewidth}
   \centerline{\begin{tabular}{|p{1cm}<{\centering}|p{1cm}<{\centering}|p{1cm}<{\centering}|p{1cm}<{\centering}|}
  \hline
  $\beta$ &0.2&0.5&0.8\\
  \hline
  OIM+IR & \textbf{50.1} & {48.0} & {46.3} \\
  \hline
  \end{tabular}}
\end{minipage}
  \end{center}
  \end{table}

\begin{table}[t]
\caption{Detection performance comparison of proposed approach on the VOC2007 with various configurations.}
\label{test_ablation}
\scriptsize
\begin{center}
\resizebox{.95\columnwidth}{12mm}{
\begin{tabular}{|p{1.4cm}<{\centering}|p{1.4cm}<{\centering}|p{1.5cm}<{\centering}|p{0.5cm}<{\centering}|p{1cm}<{\centering}|}
\hline
SG & AG & OIM(SG+AG) & IR &mAP (\%)\\
\hline
 & & & & 34.8 \\
\hline
$\surd$ & & & & 42.2 \\
\hline
 &$\surd$& & & 46.7 \\
\hline
 $\surd$& $\surd$& $\surd$ & & {48.2} \\
\hline
 & & &$\surd$ & {43.4} \\
\hline
 $\surd$& $\surd$&$\surd$ &$\surd$ & \textbf{50.1} \\
\hline
\end{tabular}}
\end{center}
\end{table}

\subsection{Ablation Experiments}
We performed ablation experiments to illustrate the effect of parameters introduced in proposed object instance mining ($\alpha$) and instance reweighted loss ($\beta$). Table 5 indicates when parameter $\alpha$ ($\alpha_1$ used in the first 70K and $\alpha_2$ used in the following 20K iterations) becomes smaller or larger, the performance of proposed approach will degrade. If the parameter $\alpha$ is too small, very less instances will be selected in the appearance graph for training. It results in that in many images, only the most discriminative object is selected and used for training. If the parameter $\alpha$ is too large, many false instances (background proposals) will be employed for training and it also leads to performance drop. For the proposed instance reweighted loss, as also can be seen from the Table 5, with the increasing of $\beta$ the performance decreases. 

We also studied the WSOD performance by only using appearance graph (AG) or spatial graph (SG) to evaluate their effectiveness separately. The first two columns in Table 6 illustrate the experimental results in terms of mAP on the VOC2007 \textit{test} set. We can see that the performance can be significantly improved for WSOD by only using appearance or spatial graph. 

\begin{figure}[t]
\centering
\includegraphics[scale = .62]{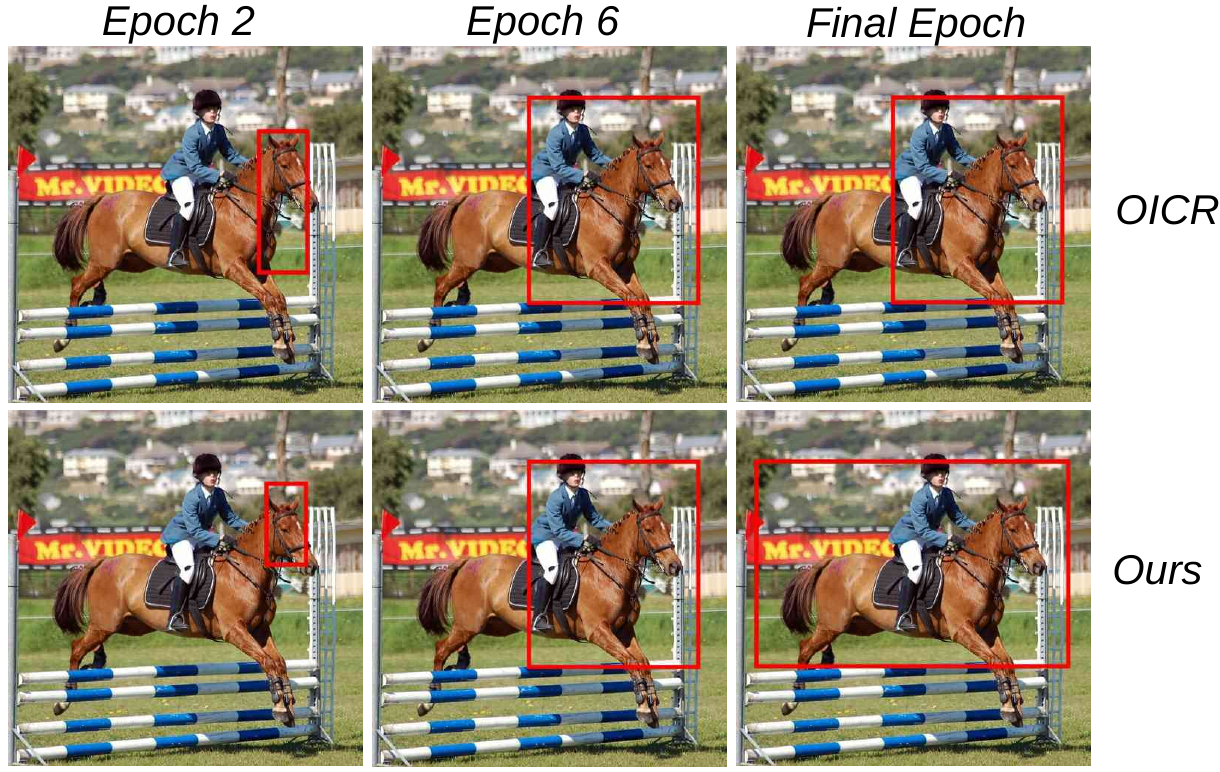}
\caption{Evolution of object detection during learning process w/o using instance reweighted loss (from left to right). The upper part of each subfigure is the result of OICR \cite{tang2017multiple} and the lower part is the result of our method.}
\label{fig-evolution2}
\end{figure}

The effectiveness of the proposed instance reweighted loss is also evaluated. We apply the network structure in OICR but just replace the loss with instance reweighted loss. The performance achieved using the proposed instance reweighted loss in terms of mAP on the VOC2007 \textit{test} set is shown in Table \ref{test_ablation}. It can be seen that the mAP can be improved from 41.2\% \cite{tang2017multiple} to 43.4\% by only using instance reweighted loss. The visual comparison shown in Figure \ref{fig-evolution2} also illustrates that larger portion of the object can be gradually detected using the proposed loss. By incorporating the OIM with instance reweighted loss, the best performance (mAP 50.1\%) can be achieved.

\section{Conclusion}
In this paper, an end-to-end object instance mining framework has been presented to address the limitations of existing approaches for WSOD. Object instance mining algorithm is performed using spatial and appearance graphs to make the network learn less discriminative object instances. Thus more possible objects belonging to the same class can be detected accordingly. Without introducing any extra count information, the proposed approach has achieved improved performance comparable to many state-of-the-art results. The object instance reweighted loss is designed to further help the OIM by learning the larger portion of the target object instances in each image. Experimental results on two publicly available datasets illustrate that the proposed approach achieves competitive or superior performance than state-of-the-art methods for WSOD. 

\bibliography{aaai_arxiv}

\begin{thebibliography}{}

\bibitem[\protect\citeauthoryear{Arun, Jawahar, and
  Kumar}{2019}]{arun2019dissimilarity}
Arun, A.; Jawahar, C.; and Kumar, M.~P.
\newblock 2019.
\newblock Dissimilarity coefficient based weakly supervised object detection.
\newblock In {\em Proceedings of the IEEE Conference on Computer Vision and
  Pattern Recognition},  9432--9441.

\bibitem[\protect\citeauthoryear{Bilen and Vedaldi}{2016}]{bilen2016weakly}
Bilen, H., and Vedaldi, A.
\newblock 2016.
\newblock Weakly supervised deep detection networks.
\newblock In {\em Proceedings of the IEEE Conference on Computer Vision and
  Pattern Recognition},  2846--2854.

\bibitem[\protect\citeauthoryear{Diba \bgroup et al\mbox.\egroup
  }{2017}]{diba2017weakly}
Diba, A.; Sharma, V.; Pazandeh, A.; Pirsiavash, H.; and Van~Gool, L.
\newblock 2017.
\newblock Weakly supervised cascaded convolutional networks.
\newblock In {\em Proceedings of the IEEE conference on computer vision and
  pattern recognition},  914--922.

\bibitem[\protect\citeauthoryear{Dietterich, Lathrop, and
  Lozano-P{\'e}rez}{1997}]{dietterich1997solving}
Dietterich, T.~G.; Lathrop, R.~H.; and Lozano-P{\'e}rez, T.
\newblock 1997.
\newblock Solving the multiple instance problem with axis-parallel rectangles.
\newblock {\em Artificial intelligence} 89(1-2):31--71.

\bibitem[\protect\citeauthoryear{Everingham \bgroup et al\mbox.\egroup
  }{2010}]{everingham2010pascal}
Everingham, M.; Van~Gool, L.; Williams, C.~K.; Winn, J.; and Zisserman, A.
\newblock 2010.
\newblock The pascal visual object classes (voc) challenge.
\newblock {\em International journal of computer vision} 88(2):303--338.

\bibitem[\protect\citeauthoryear{Everingham \bgroup et al\mbox.\egroup
  }{2015}]{everingham2015pascal}
Everingham, M.; Eslami, S.~A.; Van~Gool, L.; Williams, C.~K.; Winn, J.; and
  Zisserman, A.
\newblock 2015.
\newblock The pascal visual object classes challenge: A retrospective.
\newblock {\em International journal of computer vision} 111(1):98--136.

\bibitem[\protect\citeauthoryear{Gao \bgroup et al\mbox.\egroup
  }{2018}]{gao2018c}
Gao, M.; Li, A.; Yu, R.; Morariu, V.~I.; and Davis, L.~S.
\newblock 2018.
\newblock C-wsl: Count-guided weakly supervised localization.
\newblock In {\em Proceedings of the European Conference on Computer Vision
  (ECCV)},  152--168.

\bibitem[\protect\citeauthoryear{Girshick}{2015}]{girshick2015fast}
Girshick, R.
\newblock 2015.
\newblock Fast r-cnn.
\newblock In {\em Proceedings of the IEEE international conference on computer
  vision},  1440--1448.

\bibitem[\protect\citeauthoryear{Jie \bgroup et al\mbox.\egroup
  }{2017}]{jie2017deep}
Jie, Z.; Wei, Y.; Jin, X.; Feng, J.; and Liu, W.
\newblock 2017.
\newblock Deep self-taught learning for weakly supervised object localization.
\newblock In {\em Proceedings of the IEEE Conference on Computer Vision and
  Pattern Recognition},  1377--1385.

\bibitem[\protect\citeauthoryear{Li \bgroup et al\mbox.\egroup
  }{2019}]{li2019weakly}
Li, X.; Kan, M.; Shan, S.; and Chen, X.
\newblock 2019.
\newblock Weakly supervised object detection with segmentation collaboration.
\newblock {\em arXiv preprint arXiv:1904.00551}.

\bibitem[\protect\citeauthoryear{Liu \bgroup et al\mbox.\egroup
  }{2016}]{liu2016ssd}
Liu, W.; Anguelov, D.; Erhan, D.; Szegedy, C.; Reed, S.; Fu, C.-Y.; and Berg,
  A.~C.
\newblock 2016.
\newblock Ssd: Single shot multibox detector.
\newblock In {\em European conference on computer vision},  21--37.
\newblock Springer.

\bibitem[\protect\citeauthoryear{Oquab \bgroup et al\mbox.\egroup
  }{2015}]{oquab2015object}
Oquab, M.; Bottou, L.; Laptev, I.; and Sivic, J.
\newblock 2015.
\newblock Is object localization for free?-weakly-supervised learning with
  convolutional neural networks.
\newblock In {\em Proceedings of the IEEE Conference on Computer Vision and
  Pattern Recognition},  685--694.

\bibitem[\protect\citeauthoryear{Pan \bgroup et al\mbox.\egroup
  }{2019}]{pan2019low}
Pan, T.; Wang, B.; Ding, G.; Han, J.; and Yong, J.
\newblock 2019.
\newblock Low shot box correction forweakly supervised object detection.
\newblock In {\em Proceedings of the Twenty-Eighth International Joint
  Conference on Artificial Intelligence},  890--896.

\bibitem[\protect\citeauthoryear{Redmon \bgroup et al\mbox.\egroup
  }{2016}]{redmon2016you}
Redmon, J.; Divvala, S.; Girshick, R.; and Farhadi, A.
\newblock 2016.
\newblock You only look once: Unified, real-time object detection.
\newblock In {\em Proceedings of the IEEE conference on computer vision and
  pattern recognition},  779--788.

\bibitem[\protect\citeauthoryear{Ren \bgroup et al\mbox.\egroup
  }{2015}]{ren2015faster}
Ren, S.; He, K.; Girshick, R.; and Sun, J.
\newblock 2015.
\newblock Faster r-cnn: Towards real-time object detection with region proposal
  networks.
\newblock In {\em Advances in neural information processing systems},  91--99.

\bibitem[\protect\citeauthoryear{Russakovsky \bgroup et al\mbox.\egroup
  }{2015}]{russakovsky2015imagenet}
Russakovsky, O.; Deng, J.; Su, H.; Krause, J.; Satheesh, S.; Ma, S.; Huang, Z.;
  Karpathy, A.; Khosla, A.; Bernstein, M.; et~al.
\newblock 2015.
\newblock Imagenet large scale visual recognition challenge.
\newblock {\em International journal of computer vision} 115(3):211--252.

\bibitem[\protect\citeauthoryear{Shen \bgroup et al\mbox.\egroup
  }{2018}]{shen2018generative}
Shen, Y.; Ji, R.; Zhang, S.; Zuo, W.; and Wang, Y.
\newblock 2018.
\newblock Generative adversarial learning towards fast weakly supervised
  detection.
\newblock In {\em Proceedings of the IEEE Conference on Computer Vision and
  Pattern Recognition},  5764--5773.

\bibitem[\protect\citeauthoryear{Shen \bgroup et al\mbox.\egroup
  }{2019}]{shen2019cyclic}
Shen, Y.; Ji, R.; Wang, Y.; Wu, Y.; and Cao, L.
\newblock 2019.
\newblock Cyclic guidance for weakly supervised joint detection and
  segmentation.
\newblock In {\em Proceedings of the IEEE Conference on Computer Vision and
  Pattern Recognition},  697--707.

\bibitem[\protect\citeauthoryear{Singh and Lee}{2019}]{singh2019you}
Singh, K.~K., and Lee, Y.~J.
\newblock 2019.
\newblock You reap what you sow: Using videos to generate high precision object
  proposals for weakly-supervised object detection.
\newblock In {\em Proceedings of the IEEE Conference on Computer Vision and
  Pattern Recognition},  9414--9422.

\bibitem[\protect\citeauthoryear{Tang \bgroup et al\mbox.\egroup
  }{2017}]{tang2017multiple}
Tang, P.; Wang, X.; Bai, X.; and Liu, W.
\newblock 2017.
\newblock Multiple instance detection network with online instance classifier
  refinement.
\newblock In {\em Proceedings of the IEEE Conference on Computer Vision and
  Pattern Recognition},  2843--2851.

\bibitem[\protect\citeauthoryear{Tang \bgroup et al\mbox.\egroup
  }{2018}]{tang2018pcl}
Tang, P.; Wang, X.; Bai, S.; Shen, W.; Bai, X.; Liu, W.; and Yuille, A.~L.
\newblock 2018.
\newblock Pcl: Proposal cluster learning for weakly supervised object
  detection.
\newblock {\em IEEE transactions on pattern analysis and machine intelligence}.

\bibitem[\protect\citeauthoryear{Uijlings \bgroup et al\mbox.\egroup
  }{2013}]{uijlings2013selective}
Uijlings, J.~R.; Van De~Sande, K.~E.; Gevers, T.; and Smeulders, A.~W.
\newblock 2013.
\newblock Selective search for object recognition.
\newblock {\em International journal of computer vision} 104(2):154--171.

\bibitem[\protect\citeauthoryear{Wan \bgroup et al\mbox.\egroup
  }{2018}]{wan2018min}
Wan, F.; Wei, P.; Jiao, J.; Han, Z.; and Ye, Q.
\newblock 2018.
\newblock Min-entropy latent model for weakly supervised object detection.
\newblock In {\em Proceedings of the IEEE Conference on Computer Vision and
  Pattern Recognition},  1297--1306.

\bibitem[\protect\citeauthoryear{Wan \bgroup et al\mbox.\egroup
  }{2019}]{wan2019c}
Wan, F.; Liu, C.; Ke, W.; Ji, X.; Jiao, J.; and Ye, Q.
\newblock 2019.
\newblock C-mil: Continuation multiple instance learning for weakly supervised
  object detection.
\newblock In {\em Proceedings of the IEEE Conference on Computer Vision and
  Pattern Recognition},  2199--2208.

\bibitem[\protect\citeauthoryear{Wei \bgroup et al\mbox.\egroup
  }{2018}]{wei2018ts2c}
Wei, Y.; Shen, Z.; Cheng, B.; Shi, H.; Xiong, J.; Feng, J.; and Huang, T.
\newblock 2018.
\newblock Ts2c: Tight box mining with surrounding segmentation context for
  weakly supervised object detection.
\newblock In {\em Proceedings of the European Conference on Computer Vision
  (ECCV)},  434--450.

\bibitem[\protect\citeauthoryear{Zhang \bgroup et al\mbox.\egroup
  }{2018a}]{zhang2018zigzag}
Zhang, X.; Feng, J.; Xiong, H.; and Tian, Q.
\newblock 2018a.
\newblock Zigzag learning for weakly supervised object detection.
\newblock In {\em Proceedings of the IEEE Conference on Computer Vision and
  Pattern Recognition},  4262--4270.

\bibitem[\protect\citeauthoryear{Zhang \bgroup et al\mbox.\egroup
  }{2018b}]{zhang2018w2f}
Zhang, Y.; Bai, Y.; Ding, M.; Li, Y.; and Ghanem, B.
\newblock 2018b.
\newblock W2f: A weakly-supervised to fully-supervised framework for object
  detection.
\newblock In {\em Proceedings of the IEEE Conference on Computer Vision and
  Pattern Recognition},  928--936.

\end{thebibliography}
\bibliographystyle{aaai}
\end{document}